\documentclass[sigconf]{acmart}

\usepackage{graphicx}
\usepackage{subfigure}
\usepackage{amsmath}
\usepackage{booktabs}

\AtBeginDocument{%
  \providecommand\BibTeX{{%
    \normalfont B\kern-0.5em{\scshape i\kern-0.25em b}\kern-0.8em\TeX}}}

\begin{document}

\title{From Extreme Multi-label to Multi-class: A Hierarchical Approach for Automated ICD-10 Coding Using Phrase-level Attention}

\author{Cansu Sen}
\email{cansu@codametrix.com}
\affiliation{%
  \institution{CodaMetrix}
}

\author{Bingyang Ye}
\email{bingyang@codametrix.com}
\affiliation{%
  \institution{CodaMetrix}
}

\author{Javed Aslam}
\email{jay@codametrix.com}
\affiliation{%
  \institution{CodaMetrix}
}

\author{Amir Tahmasebi}
\email{atahmasebi@enlitic.com}
\affiliation{%
  \institution{CodaMetrix, Enlitic}
}


\begin{abstract}
Clinical coding is the task of assigning a set of alphanumeric codes, referred to as ICD (International Classification of Diseases), to a medical event based on the context captured in a clinical narrative. The latest version of ICD, ICD-10, includes more than 70,000 codes. As this is a labor-intensive and error-prone task, automatic ICD coding of medical reports using machine learning has gained significant interest in the last decade. Existing literature has modeled this problem as a multi-label task. Nevertheless, such multi-label approach is challenging due to the extremely large label set size. Furthermore, the interpretability of the predictions is essential for the end-users (e.g., healthcare providers and insurance companies). In this paper, we propose a novel approach for automatic ICD coding by reformulating the extreme multi-label problem into a simpler multi-class problem using a hierarchical solution. We made this approach viable through extensive data collection to acquire phrase-level human coder annotations to supervise our models on learning the specific relations between the input text and predicted ICD codes. Our approach employs two independently trained networks, the sentence tagger and the ICD classifier, stacked hierarchically to predict a codeset for a medical report. The sentence tagger identifies \emph{focus sentences} containing a medical event or concept relevant to an ICD coding. Using a supervised attention mechanism, the ICD classifier then assigns each focus sentence with an ICD code. The proposed approach outperforms strong baselines by large margins of 23\% in subset accuracy, 18\% in micro-F1, and 15\% in instance-based F-1. With our proposed approach, interpretability is achieved not through implicitly learned attention scores but by attributing each prediction to a particular sentence and words selected by human coders. 
\end{abstract}



\maketitle

\section{Introduction}
Clinical notes are free text narratives generated by healthcare professionals during patient encounters. These notes often contain crucial details regarding patient health, along with expert insights. They are typically accompanied by a set of alphanumeric codes from the International Classification of Diseases (ICD), which present a standardized way of capturing and communicating diagnoses and procedures performed during the encounter within care workflows. ICD codes have various uses, ranging from billing to predictive modeling of patient state \cite{choi2016doctor, avati2018improving, mullenbach2018explainable} and population health analytics. Typically, there is a dedicated team of trained personnel within the billing department who assign ICD codes for every encounter based on the content of corresponding clinical notes. Because manual code assignment, referred to as \textit{coding}, is time-consuming and prone to human error, automatic coding has gained vast interest among machine learning researchers in recent years \cite{xu2019multimodal, vu2020label, shi2017towards}.

Automatic ICD coding is a challenging task for several reasons. First, the label space is high-dimensional with over 70,000 codes according to 2020 ICD-10-CM guidelines \footnote{https://www.cms.gov/Medicare/Coding/ICD10/2020-ICD-10-CM}. When modeled as a machine learning task, this corresponds to an extreme multi-label prediction. 
Second, clinical notes contain a summary of all findings and diagnoses observed during the current visit, as well as relevant findings from prior visits for comparison. For billing ICD coding, certain findings are only ``codable" that are essential for the current visit. For example, R91.8 (i.e., Other nonspecific abnormal finding of lung field) is a relevant ICD-10 code to capture and code a lung nodule detected during a follow-up chest CT imaging procedure; however, it is not an appropriate code if the nodule was detected incidentally during a chest XRAY imaging performed for a fracture diagnosis. Thus, identifying relevant information for ICD coding is challenging for machine learning-based approaches. This has resulted in ICD \textit{over-coding}, predicting too many irrelevant ICD codes.
Finally, neural network-based text classification, which is frequently used for automatic ICD coding, tends to be ``black-box'' in the sense that the models are difficult to interpret. Nevertheless, in the clinical setting, interpretability is paramount to provide clinicians insights into how models make particular predictions.

Many works have recently tried to (partially) address these challenges and propose methods for automatic ICD coding of clinical notes. Since each document may be assigned one or multiple ICD codes, existing work heavily leans toward multi-label modeling  \cite{shi2017towards,mullenbach2018explainable, vu2020label}.
Even though this is an intuitive approach, it struggles with resolving some of the challenges, such as large label space. The task becomes predicting a handful of labels among tens of thousands of possible label candidates. These works also input either the entire note or only a portion of the note into the machine learning model. The former approach complicates the problem due to extreme input size, while the latter approach risks ignoring potentially relevant information from other note sections. Further, many works attempt to seek a remedy for the lack of interpretability of the neural network models by using attention mechanisms in their model design. However, recent research has shown that attention scores may not necessarily give selective and meaningful scores for the input words \cite{sen2020human, jain2019attention}. Overall, these approaches remain suboptimal. 

This paper embraces a novel approach for automatic ICD coding by turning the extreme multi-label task into a multi-class classification problem with a hierarchical solution. Our proposed architecture consists of two standalone models working at different levels. The first level model, referred to as the \textit{sentence tagger}, takes all sentences from a clinical report as input and identifies the ones that contain a clinical event or concept relevant to ICD coding.
Using a Long Short-Term Memory Network (LSTM) architecture, it tags each sentence either as a focus sentence or not. If a sentence is tagged as a focus sentence, it is then passed to the second level model, the \textit{ICD classifier}. The ICD classifier consists of multiple LSTM networks and a supervised attention mechanism for assigning each focus sentence an appropriate ICD code. Finally, we implement an end-to-end pipeline using our two models hierarchically to generate the final codeset for the whole note. 

Our approach relies on sentence-level supervision for the first level to identify focus sentences and word-level supervision for the second level with attention supervision. To this end, we collected a large dataset, PAP: \textbf{P}hrase-level coder \textbf{A}nnotation for \textbf{P}athology, from a large medical institution. We recruited multiple coders for ICD coding of pathology reports and annotated the corresponding evidence for each assigned ICD code. 


The biggest barrier for the deep learning-based models proposed for automatic ICD coding is their lack of interpretability. Our model solves this problem in two ways. First, since our sentence tagger operates at the sentence level, we can point at a single sentence as the reason for a particular ICD prediction. Second, the supervised attention mechanism further identifies the exact word(s) within a sentence as the prediction reasoning. Supervising the attention mechanism with human feedback ensures learning from human coders instead of learning the attention scores unsupervised.

Our contributions in this paper are as follows: 
\begin{itemize}
    \item To the best of our knowledge, we propose the first work to model the ICD prediction task as a multi-class classification problem by breaking it into two sub-tasks. We achieve this through an extensive data collection effort.
    \item To the best of our knowledge, our proposed method is the first in its kind to address interpretability for automatic ICD coding via rendering superior interpretability with the help of two-level design and supervised attention. Interpretability is vital for the ICD prediction task for bringing it one step closer to full automation. 
    \item Our extensive experimental results demonstrate significant outperformance compared to existing state-of-the-art baseline models on a relatively large dataset. 
\end{itemize}

\section{Related Work}
In recent years, several works have proposed machine learning models for automatic ICD coding. Many of these approaches rely on variants of Convolutional Neural Networks (CNNs) and LSTMs. In \cite{xu2019multimodal}, authors use an ensemble of a character level CNN, Bidirectional-LSTM, and word level CNN to derive ICD code predictions. Another study proposes a tree-of-sequences LSTM architecture to capture the hierarchical relationship among codes and the semantics of each code  \cite{xie2018neural}.

Other works further incorporate the attention mechanisms introduced by \cite{bahdanau2014neural} to better utilize information buried in longer input sequences. In \cite{mullenbach2018explainable}, authors explore attentional convolutional networks. They aim to bring interpretability to a model's predicted codes with the help of per-label attention. In \citet{baumel2018multi}, the authors introduce a Hierarchical Attention bidirectional Gated Recurrent Unit(HA-GRU) architecture. Character-aware LSTM's with attention are utilized to generate sentence representations from specific subsections of discharge summaries \cite{shi2017towards}. In \cite{prakash2017condensed}, memory networks draw from discharge summaries and general-text from Wikipedia to predict top ICD codes. 

A major motivation factor for using attention mechanisms is to incorporate interpretability into designed architecture. However, this claim is disputed by the recent literature  \cite{sen2020human, jain2019attention}. Even though attention mechanisms help improve the model performance, attention scores may not necessarily give selective and meaningful scores for the model inputs. To overcome this challenge, ways of attention supervision have been proposed for several tasks such as image classification and event detection \cite{zhao2018document, nguyen2018killed, liu2017exploiting, nguyen2018deep}. However, supervised attention has not been applied to the ICD coding problem. 

Clinical notes are typically long documents with thousands of words in each note. ML architectures designed for ICD coding must be able to tackle this long input size. Some existing work bypasses this problem by manually extracting the most-relevant subsection from the clinical report (i.e., discharge diagnosis section from discharge summaries) \cite{shi2017towards,prakash2017condensed}. This approach simplifies the task by shortening the input length but risks discarding relevant information. Our two-level approach is capable of handling long documents, and thus, it eliminates the need to omit information or sections from the clinical notes.

\begin{figure*}
    \centering
    \includegraphics[width=0.8\textwidth]{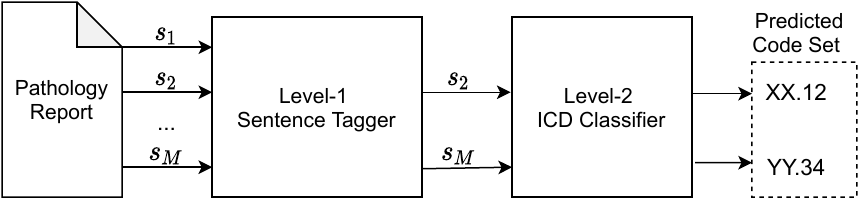}
    \caption{The proposed end-to-end hierarchical prediction pipeline.}
    \label{fig:pipeline}
\end{figure*}

A significant number of prior work, if not all, use MIMIC-III dataset \cite{johnson2016mimic} for their experimental evaluation \cite{mullenbach2018explainable, xie2018neural, shi2017towards,li2018automated,huang2019empirical,xu2019multimodal}. The MIMIC-III dataset contains ICD-9 codes instead of ICD-10, thus naturally limiting the label set size down to a few thousand. Some works further limit the label space to the most frequent codes \cite{shi2017towards,xu2019multimodal}. Our experimental evaluation is based on the latest version of ICD-10-CM (2020).


\section{Methods}

\begin{figure}
    \centering
    \includegraphics[width=1.0\columnwidth]{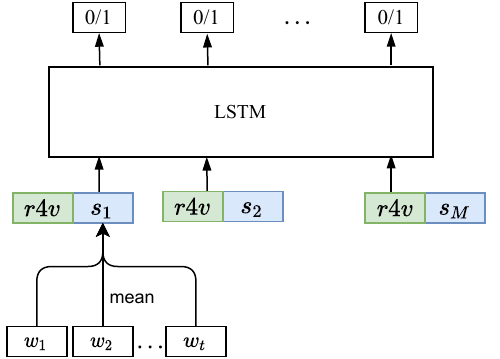}
    \caption{Level one architecture: Sentence Tagger.}
    \label{fig:level-1}
\end{figure}

\begin{figure*}
    \centering
    \includegraphics[width=0.8\textwidth]{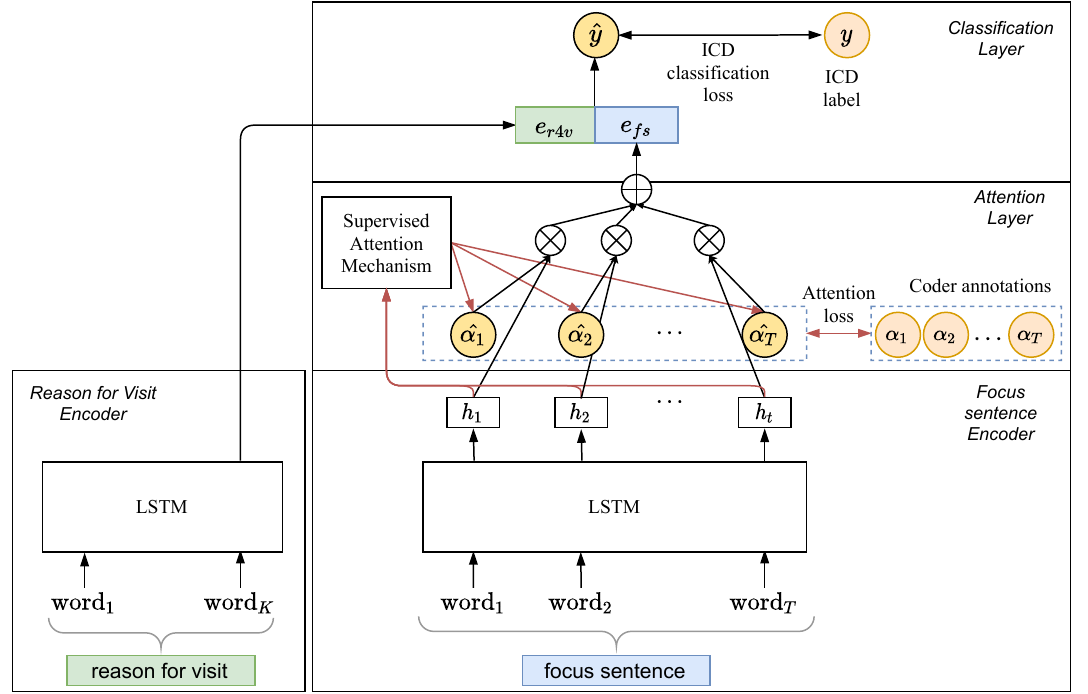}
    \caption{Level two architecture: ICD Classifier.}
    \label{fig:level-2}
\end{figure*}

\subsection{Problem Definition}
Let $\mathcal{L}$ represents the ICD-10 label set. We are given a set of $N$ clinical reports $\mathcal{D}=\{\mathcal{D}^1, \dots, \mathcal{D}^N\}$. An ICD codeset $\mathcal{L}^i \subset \mathcal{L}$ is assigned to each report $\mathcal{D}^i$ by human coders. 
 
 An ICD codeset $\mathcal{L}^i \subset \mathcal{L}$ is assigned to each report $\mathcal{D}^i$ by human coders. ICD prediction task is to learn a parameterized function $f_\theta(\cdot)$ that maps $\mathcal{D}^i \to \mathcal{L}^i$, generalizing to unseen instances.

Most existing works model this problem as a multi-label classification task \cite{xu2019multimodal,mullenbach2018explainable,shi2017towards,baumel2018multi}. Thus, they predict the probability $p^i_l \in \{0,1\}$ for all $l \in \mathcal{L}$.

In this paper, we instead take a two-step hierarchical approach. We first predict which sentences within a report contain a medical event or concept that usually triggers an ICD code assignment by human coders. After identifying these relevant sentences, referred to as \emph{focus sentences}, we then predict the exact ICD code for each focus sentence. Finally, all predicted codes are assembled into a final codeset corresponding to the whole note.

Henceforth dropping the superscript $i$ for simplicity, given a clinical report $\mathcal{D}$ consisting of $M$ sentences $s= [s_{1}, \dots, s_{M}]$, we formally define the ICD prediction problem as follows: 

1) identify $K$ focus sentences $f \subset s$ through a function mapping 
$s_{1:M} \to y_{1:M} \in (0,1)$, sentences that are mapped to 1 being the focus sentences,

2) for each focus sentence $f_i$, assign an ICD code through a function mapping  $f_{i} \to y \in \mathcal{L}$.

This approach allows us to formulate the problem as a binary classification task in the first step $s_i \to (0,1)$ and a multi-class classification task in the second step $s_i \to c_i$, $c_i \in \mathcal{L}$. Final predicted codeset for the report $\mathcal{D}$ becomes the union of all predicted ICD codes for the focus sentences. 


\subsection{Proposed Hierarchical Approach}
The proposed pipeline is demonstrated in Figure \ref{fig:pipeline}. The sentence tagger consists of an LSTM network, which tags each sentence either as a focus sentence or not. If a sentence is tagged as a focus sentence, it is then passed into the second level model, the ICD classifier. 

The ICD classifier consists of multiple LSTM networks and a supervised attention mechanism for assigning an ICD code to each focus sentence. We utilize a human curated dataset of pathology notes labeled by multiple coders. This dataset contains word-level annotations from the human coders when performing the ICD assignment task. Hence, collected annotations demonstrate which words/phrases made the coders assign a particular ICD code. This information is used for model training in the form of attention supervision. 

In addition to the clinical note's body (free-form text), we include a second input, referred to as \emph{``reason for visit''}. As the name implies, \textit{Reason for visit} is a structured meta-data, a free-form text capturing the reason for the clinical event (e.g., shave biopsy, upper eyelid lesion).

The following sections describe the model details of each level.

\subsection{Level 1: Sentence Tagger}

As shown in Figure \ref{fig:level-1}, the input to the sentence tagger consists of the sentences from a report and the reason for visit sentence. We encode sentences as following: We calculate the coordinate-wise mean word embedding $e_i$ for all words in the sentence $w_1, \dots, w_T$. We use the same process to compute an encoded representation for reason for visit $r_i$ using all words in reason for visit. We then concatenate these two vectors to compute the final sentence level representation: 

\begin{equation}
    s_i = [e_i, r_i]
\end{equation}

Let our input sentence representations be $s_1,…,s_M$,  $\mathcal{T} \in \{0,1\} $ be our tag set, and $y_i$ be the tag of sentence $s_i$. Our goal is to predict an output sequence $\hat{y}_1, \dots, \hat{y}_M$ where $\hat{y}_i \in \mathcal{T}$. To compute  $\hat{y}$, we first pass the derived sentence representations through an LSTM network as follows: 
\begin{equation}
h_{i} = \text{LSTM}_{st}(s_i, h_{i-1}) 
\end{equation}

Then our prediction for $\hat{y}$ is:

\begin{equation}
    \hat{y}_i = \text{argmax}_j ( \text{log} \text{Softmax}(Wh_i+b))j
\end{equation}

We use negative log-likelihood loss to train the sentence tagger.

For realizing the sentence tagger, we opt for a simpler model where sentences are represented as average word embedding instead of training a hierarchical LSTM network, one for words and another for sentences. This decision is based on the experimental observation that the more complex model taking too much time while providing a relatively small advantage.

\subsection{Level 2: ICD Classifier}
The ICD classifier consists of two LSTM networks and a supervised attention mechanism.  We model the ICD prediction task as a multi-class classification problem. Thus, the ICD classifier predicts a single ICD code for each input. The architecture is presented in Figure \ref{fig:level-2}.

\subsubsection{Sentence encoder}
Our ICD classifier first utilizes an encoding layer to map words into real-valued vector representations where semantically-similar words are mapped close to one another. We use a pre-trained word embedding set $\phi$ for this mapping: $x_{it} = \phi w_{it}$. We then employ a recurrent layer to embed vector representations of words into hidden states, processing words one at a time. 

We use the same process to encode both the focus sentences and reason for visit. That is, we pass the predicted focus sentence and the relevant reason for visit through two independent LSTM networks. 

The encoded reason for visit $e_{r4v}$ is the output of the last step of $\text{LSTM}_{r4v}$.  
Then, assuming that $x_t$ is the word embedding for $t$-th word from the focus sentence $s$, we compute:

\begin{align}
h_{t} &= \text{LSTM}_{fs}(x_t, h_{t-1}) \\ 
u_{t} &= \Phi(W h_{t} + b) \\
\hat{\alpha}_{t} &= \frac{\exp(u_{t}^\top h_{T})}{\sum_t \exp(u_{t}^\top h_{T})}
\end{align}

where $W$ and $b$ are trainable parameters,  $h_t$ is the hidden state, $\Phi$ is the hyperbolic tangent function, and $\hat{\alpha}_t$ are the attention scores for each word, computed through a softmax function. Resulting attention scores $[\hat{\alpha}_{1}, \dots, \hat{\alpha}_{T}]$ are utilized by the subsequent layer.

\subsubsection{Supervising the attention mechanism}
The next layer is designed to supervise the attention mechanism to drive the learned attention scores closer to that of human coders. Attention supervision has recently been used with a similar purpose by some researchers \cite{zhao2018document, nguyen2018killed, liu2017exploiting}. We use a similar attention supervision method to force our model to pay more attention to the words/phrases selected by human annotators. 

To this end, we take the squared error as the general loss of attention at the word level. 

\begin{equation}
    J_a = \sum_{t=1}^T ( \hat{\alpha}_{t} - \alpha_{t})^{2}
\end{equation}

where $\hat{\alpha}_{t}$ is the predicted attention score by the sentence encoder layer and $\alpha_{t}$ is the ground truth attention score assigned to that word by the human coders. We represent $\alpha$ as a binary vector where 1 denotes a word receiving attention from the coder. 

This objective optimizes the model to allocate correct importance scores to every word. By providing word-level supervision to the ICD classification model, we are able to teach it to focus on the most relevant areas selected by humans and thereby improve the quality of sentence representations along with the overall performance.

\subsubsection{Classification layer}
The final layer of this model is the classification layer. In this layer, we first compute a context vector $c$ as follows: 

\begin{align}
e_{fs} &= \sum_t \hat{\alpha}_{t} h_{t}. \\
c &= [e_{fs}, e_{r4v}]
\end{align}

Here, $e_{fs}$ is the weighted sum of hidden representation of words by the supervised attention scores, and it can be seen as a dense summary of the focus sentence. The context vector $c$ is the concatenation of $e_{fs}$ and $e_{r4v}$. The classification layer uses $c$ and assigns a probability to each possible class. We use the cross-entropy loss as the classification objective function where $\hat{y}$ is the prediction, and $y$ is the ground truth label.

\begin{equation}
    J_c = - (y \log(\hat{y})+(1-y)\log(1-\hat{y})).
\end{equation}

Our model utilizes a joint training paradigm for the sentence encoder and attention supervision. Thus, we define a joint loss function in the training process upon the losses specified for different subparts as follows:
\begin{equation}
\label{eqn:joint_loss}
    J(\theta) = \sum(J_c +  \lambda J_a)
\end{equation}
where $\theta$ denotes, as a whole, the parameters used in our model, and $\lambda$ is the hyper-parameter for striking a balance among ICD classification and attention supervision. By integrating these two objectives similar to the multi-task setting, the two tasks aid each other's training.

\subsection{Inference with our Hierarchical Model}
We train the sentence tagger and the ICD classifier separately as we have the ground truth at each level. After training each model individually, these trained models are employed end-to-end in a hierarchical manner at the inference phase. After the sentence tagger decides which sentences should be considered further, the ICD classifier takes in each of those sentences and assigns an ICD code to them. The final predicted codeset for the entire clinical report then becomes the union of all predicted ICD codes.

\section{Experimental Results}

\subsection{PAP Dataset}
PAP dataset is a collection of $39,000$ pathology reports from a large medical institution. We asked four ICD coders (three coders and one senior coder) to code pathology reports and annotate the phrases within the report that trigger a particular ICD code assignment. The coders annotate $ \sim66,000$ phrases. 

Each note is annotated by two coders randomly selected from the pool of three. The annotations from the two coders are merged as follows:
\begin{enumerate}
    \item If the annotated text from both coders match exactly OR partially overlap, the union of the two annotations is considered as the final ground truth annotation.
    \item If the annotated text from both coders do not overlap, the case is further reviewed by the senior coder to mark the final annotation.
\end{enumerate}

We use BRAT (https://brat.nlplab.org/) as annotation tool. We report inter-annotator agreement (IAA) scores using Cohen’s kappa in Table \ref{tab:iaa}. According to Table \ref{tab:iaa}, the inter-annotator agreement for every pair of coders is determined as substantial.

\begin{table}[]
    \centering
    \begin{tabular}{ll}
    \toprule
         Coder pair	&  IAA \\
         \midrule
        Coder-1, Coder-2 & 0.69  \\
        Coder-1, Coder-3 & 0.61 \\
        Coder-2, Coder-3 & 0.62 \\
        Coder-1, Coder-2, Coder-3  & 0.64 \\
        \midrule
        Senior-coder pair &	IAA \\
        \midrule
        Senior-coder, Coder-1 & 0.62 \\
        Senior-coder, Coder-2 & 0.61 \\
        Senior-coder, Coder-3 & 0.61 \\
        \bottomrule
    \end{tabular}
    \caption{Inter-annotator agreement (IAA) scores among every pair of coders using Cohen’s kappa.}
    \label{tab:iaa}
\end{table}

\begin{table*}[]
    \centering
    \resizebox{\textwidth}{!}{
    \begin{tabular}{@{}llll@{}}
\toprule
Focus Sentence                                                                                                & \begin{tabular}[c]{@{}l@{}}Annotated words \\ in this sentence\end{tabular} & Reason for visit                                                                                & Assigned ICD code \\ \midrule
\begin{tabular}[c]{@{}l@{}}['lung', 'adenocarcinoma', 'metastatic',\\  'to', 'lymph', 'node']\end{tabular}    & [1, 1, 0, 0, 0, 0]                                                          & \begin{tabular}[c]{@{}l@{}}['molecular', 'test', \\ 'for', 'met']\end{tabular}                  & C34.90            \\
\begin{tabular}[c]{@{}l@{}}['transformation', 'zone', 'with', \\ 'chronic', 'inflammation', '.']\end{tabular} & [0, 0, 0, 1, 1, 0]                                                          & \begin{tabular}[c]{@{}l@{}}['cervix', 'biopsy', '-', \\ 'endocervix', 'curettage']\end{tabular} & N72               \\ \bottomrule
\end{tabular}
    }
    \caption{Example of the training data for the ICD classifier. Each training instance consists of a focus sentence, attention signals corresponding to the annotated words in the focus sentence, and reason for visit. Ground truth is the ICD label assigned by the coder.}
    \label{tab:data-example}
\end{table*}

\subsubsection{Data preparation}
We first split our data into training (~70\%), validation (~15\%), and test (~15\%) sets. These sets contain $28,697$, $5,217$, and $5,118$ pathology reports, respectively. Models are trained on the reports from the training set, and the validation set is used for hyper-parameter selection. All numbers reported in this section are based on the reports from the test set. We use the same split for training all models. 

Both models use sentence-level input. Thus, we first split the reports into sentences using ScispaCy's \emph{en\_core\_sci\_lg} model which is trained on biomedical data \cite{neumann2019scispacy}. Sentences containing annotated phrases by coders are labeled as focus sentences, while all the other sentences within a note are labeled as negative (non-focus) sentences. These labels are consumed as the ground truth for training the sentence tagger model.

ICD classifier is trained using \emph{only} the focus sentences, as these are the sentences that require an ICD code assignment. Our ICD classifier is trained on a sentence-level dataset, as opposed to the training of the sentence tagger, which is performed using the report-level dataset. 

The sentence-level dataset contains four pieces of information: 

1) The focus sentence: This corresponds to the sentence selected from the report.

2) Attention supervision signals: This is a binary vector generated using the phrase-level annotations from the PAP.  ``1'' denotes a word receiving attention from the coder.

3) Reason for visit:  It is a sentence indicating the reason for the biopsy, exam, or complaint for the encounter where the pathology report was created. Coders can assign different ICD codes to a report depending on the reason for visit.

4) Ground truth label: ICD-10 code assigned to the report by the PAP coder. 

An example is provided in Table \ref{tab:data-example}.

In some rare cases, multiple ICD codes are assigned to a sentence (3\% of our dataset). We remove those cases during the model training of the ICD Classifier to prevent noise as there would be multiple ground-truth labels for the same input sentence. However, clinical reports with such sentences are still included in our end-to-end test set.

\subsubsection{Label space}
956 unique ICD codes exist in the PAP dataset. We include the labels that occur at least ten times in our training dataset and exclude the others. The resulting label set size is 410. 

\begin{figure}
    \centering
    \includegraphics[width=0.95\columnwidth]{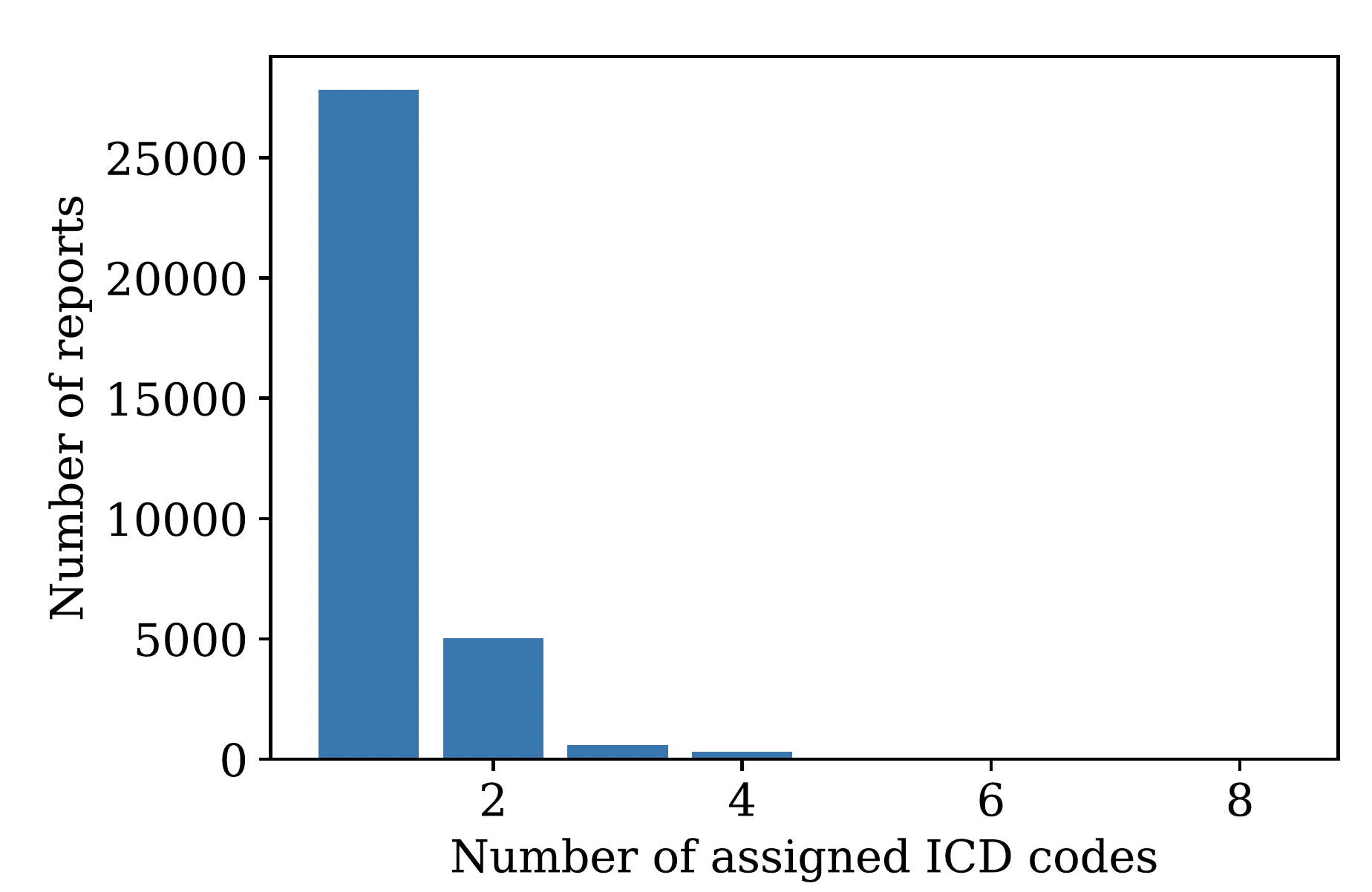}
    \caption{Histogram of ICD-10 codeset size in PAP dataset.}
    \label{fig:icd-hist}
\end{figure}

Most pathology reports are assigned one or two ICD codes. Rarely, some notes are assigned a larger codeset, up to eight codes. Figure \ref{fig:icd-hist} shows the histogram of the number of labels (ICD-10 codes) assigned per report. 

\subsection{Experimental Settings}

\textbf{Implementation Details.} We use PyTorch \cite{paszke2017automatic} for implementing all algorithms. We use Biomed embeddings \cite{moen2013distributional} for encoding words into vector representations in both models. The RNN hidden layer size is 256 for both the sentence tagger and the ICD classifier, and the attention layer size is 128. We use Adam optimizer \cite{kingma2014adam} with a learning rate of $10e-3$. We use a batch size of 32, and we train the sentence tagger for 100 epochs and the ICD classifier for 30 epochs.  

\textbf{Evaluation metrics.} We use precision, recall, and F-1 for measuring model performances. The sentence tagger is a binary classifier trained on imbalanced data, whereas the ICD classifier is a multi-class classifier.
We report macro and weighted averages for each metric. The weighted metrics calculate the score for each class independently. Then it adds them together using a weight that depends on the number of true labels of each class. The micro metrics (or accuracy) utilizes the global numbers of true positives without respect to the class. Finally, the macro average calculates the score for each class independently. However, it does not use weights while aggregating them.

For our hierarchical pipeline, the end goal is a multi-task problem. Appropriately, 
we report the subset accuracy as well as instance and micro weighted precision, recall, and F-1 values. 
The subset accuracy, also referred to as the exact match ratio, indicates the percentage of samples that have all their labels classified correctly.
While a common metric for multi-task problems, the subset accuracy ignores partially correct matches. Instance-based metrics (F-1, precision, and recall) are calculated for each label. Then their average is weighted by support - which is the number of true instances for each label.

\begin{table}[]
    \centering
    \resizebox{0.90\columnwidth}{!}{
    \begin{tabular}{lccc}
    \toprule
    & Precision & Recall & F-1 \\
    \midrule
    Micro avg. & 0.99 & 0.99 & 0.99 \\
    Macro avg. & 0.78 & 0.97 & 0.85 \\
    Weighted avg. & 0.99 & 0.99 & 0.99 \\

    \bottomrule
    \end{tabular}
    }
    \caption{Performance of the Sentence tagger on test set.}
    \label{tab:level-1}
\end{table}
\begin{table*}[]
\resizebox{0.94\textwidth}{!}{
\begin{tabular}{lccccc}
\toprule
                   & \multicolumn{1}{c}{Pooling} & \multicolumn{1}{c}{Attention} & \multicolumn{1}{c}{Supervised Attention} & \multicolumn{1}{c}{\begin{tabular}[c]{@{}c@{}}Attention \\ \& R4V\end{tabular}} & \multicolumn{1}{c}{\begin{tabular}[c]{@{}c@{}}Supervised Attention \\ \& R4V\end{tabular}} \\ 
\midrule
Accuracy              & 0.75                        & 0.76                          & 0.77                                     & 0.86                     & \textbf{0.89}                            \\

Macro Precision    & 0.66                        & 0.66                          & 0.67                                     & 0.81                    & \textbf{0.85}                            \\

Macro Recall       & 0.62                        & 0.63                          & 0.64                                     & 0.79                        & \textbf{0.83}                         \\

Macro F1           & 0.61                        & 0.62                          & 0.63                                     & 0.78                     & \textbf{0.82}                           \\

Weighted Precision & 0.77                        & 0.75                          & 0.78                                     & 0.87                     & \textbf{0.90}                            \\

Weighted Recall    & 0.75                        & 0.74                          & 0.75                                     & 0.86                       & \textbf{0.89}                         \\

Weighted F1        & 0.74                         & 0.73                          & 0.75                                      & 0.85                  & \textbf{0.89}                            \\ 
\bottomrule
\end{tabular}
}
\caption{Performance of the ICD classifier on test set.}
\label{table:icd-results}
\end{table*}
\begin{table*}[]
\resizebox{0.94\textwidth}{!}{
\begin{tabular}{lccccccc}
\toprule
& \multicolumn{1}{l}{Set Accuracy} &
\multicolumn{3}{c}{Micro}    &
\multicolumn{3}{c}{Instance}    \\ 
\cmidrule(lr){3-5}
\cmidrule(lr){6-8}  
& & F-1  & Precision & Recall & F-1   & Precision  & Recall  \\ \midrule
SVM Binary Relevance Model & 0.41  & 0.52 & 0.44      & 0.62   & 0.56  & 0.53       & 0.62    \\
LR Binary Relevance Model  & 0.59  & 0.65 & 0.61      & \textbf{0.71}   & 0.68  & 0.67       & 0.71    \\
Proposed Method    & \textbf{0.64} & \textbf{0.70} &  \textbf{0.71}      & 0.69   & \textbf{0.71}  & \textbf{0.72}       & \textbf{0.71}    \\
\bottomrule
\end{tabular}
}
\caption{Performance of the proposed end-to-end hierarchical classifier on test set.}
\label{table:end-to-end}
\end{table*}

\subsection{Performance of the Sentence Tagger}
Performance of the sentence tagger on the test set is presented in Table \ref{tab:level-1}. The sentence tagger achieves $0.99$ F-1 score and $0.85$ macro F-1 score. 

Since only one or a few sentences are considered the focus sentences from each report, the resulting dataset is largely imbalanced. We tune the \emph{weight} hyper-parameter of the \emph{torch.nn.NLLLoss} function to account for the disproportionate class sizes. The \emph{weight} hyper-parameter is a manual rescaling factor given to each class. Macro averaged precision and recall values provide further information regarding how well the model handles the imbalanced classes. A high recall value ($0.97$) indicates that most of the actual focus sentences are predicted by the model correctly. A relatively lower precision score ($0.78$) is a sign that some of the non-focus sentences are mispredicted as focus sentences, which could lead to passing too many sentences to the second level by the model.

\subsection{Performance of the ICD Classifier}
We compare the ICD Classifier with the following baseline models:
\begin{itemize}
    \item \textbf{Mean-Max Pooling Model}: This baseline model contains an LSTM network similar to our proposed model. The output of the LSTM is passed through mean and max pooling operations instead of an attention mechanism. 
    \item \textbf{Vanilla Attention Model}: This baseline model utilizes an LSTM network followed by additive attention as described in Section 3.4. However, the attention mechanism is not supervised with annotator feedback. Instead, it is learned unsupervised as commonly done in the literature \cite{bahdanau2014neural, shi2017towards, yang2016hierarchical, mullenbach2018explainable}.
    \item \textbf{Supervised Attention Model}: This is the proposed architecture. 
\end{itemize}

To see the effect of employing \emph{reason for visit}, we run the \emph{vanilla attention} and \emph{supervised attention} base architectures with and without adding the \emph{reason for visit}. For the models that reason for visit is not used, the encoded representation of the focus sentence is only propagated into the classification layer (refer to Section 3.4 for details). 
Experimental results are presented in Table \ref{table:icd-results}. The proposed architecture achieves superior performance in all metrics compared to the baseline models. We also observe that including reason for visit improves the performance drastically for both vanilla attention and supervised attention models. Among the three base architectures (i.e., pooling, vanilla attention, and supervised attention), supervised attention achieves a minor performance improvement over the other two models. However, its foremost benefit comes from the improved interpretability. 

The ICD Classifier model has a hyper-parameter, $\lambda$, that requires fine-tuning. Higher  $\lambda$ values place a higher emphasis on the attention supervision and drive the attention scores more similar to human coders. We experiment with $\lambda=[0.1, 1, 10, 100, 1000]$ values and empirically find that $\lambda=100$ performs best in our experiments. Results presented in Table  \ref{table:icd-results} is obtained with this setting.


\begin{figure}
    \centering
    \includegraphics[width=1\columnwidth]{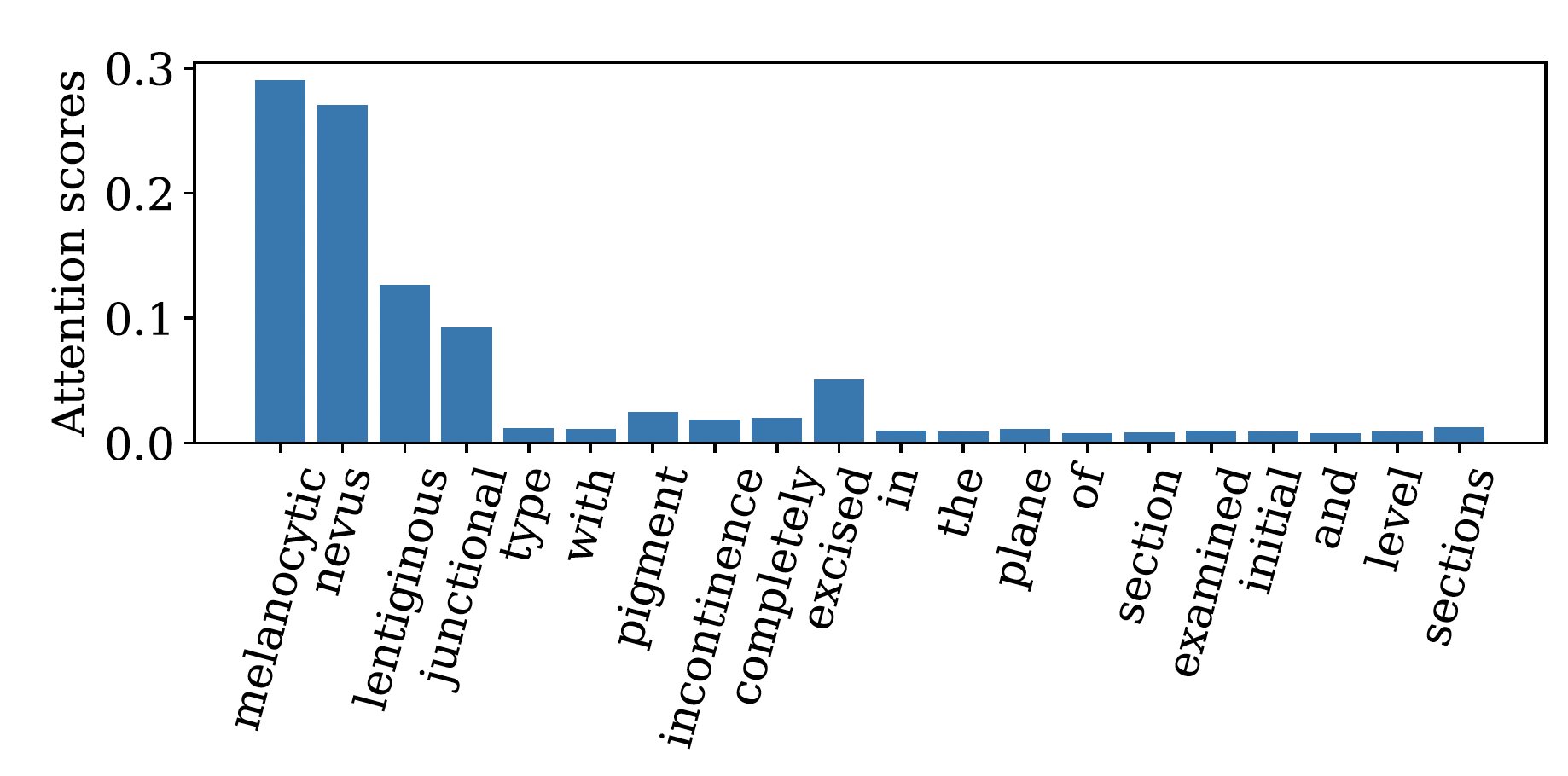}
    \vspace{-20pt}
    \caption{An example focus sentence and the attention distribution for this sentences' words. The two highest attention scores are received by the words ``melanocytic'' and ``nevus'', which are also selected by the human coders. The assigned ICD code is D22.71.}
    \label{fig:case_study_word}
\end{figure}

\subsection{End-to-end Performance Evaluation}

We compare the performance of our hierarchical pipeline with the following baseline models:
\begin{itemize}
    \item \textbf{LR Binary Relevance Model}: Binary relevance models work by decomposing the multi-label learning task into several independent binary learning tasks, one per class label \cite{luaces2012binary, zhang2018binary, tanaka2015multi}. The logistic regression model consists of $\mathcal{L}$ binary one-vs-rest classifiers for all labels. Input features are n-grams (uni-gram to tri-gram) extracted from the report. This approach is commonly used as a baseline for automatic ICD coding \cite{mullenbach2018explainable} and comparable to our model since n-gram based input format does not restrict the input size as BERT-based models do. 
    \item \textbf{SVM Binary Relevance Model}: Similar to the LR Binary Relevance Model, the SVM binary relevance model also trains a binary SVM classifier for each label in the label space.  
\end{itemize}

Experimental results are presented in Table \ref{table:end-to-end}. Our proposed pipeline outperforms all baselines by large margins.

In some cases, the sentence tagger identifies an incorrect sentence to be the focus sentence. However, these sentences may still be assigned correct ICD codes by the ICD classifier. We observe this type of behavior when there are multiple similar sentences in the report. Since the union of the predicted ICD codes is the final predicted codeset, these cases are considered correct classification by the model and do not impact the performance.

All three models we compare tend to overpredict, i.e., they predict more ICD codes than the ground-truth codeset. However, our model suffers the least from this problem. Our model can further address this problem by fine-tuning the sentence tagger to transfer more or fewer sentences to the ICD classifier. 

By design, the sentence tagger may predict all sentences from a report as non-focus. This is an undesired event since all reports receive at least one ICD code. In rare situations where we observe this phenomenon, we label the sentence with the highest probability as the focus sentence. Since 72\% of the reports in our dataset have a single ICD code, labeling one sentence as the focus is the most fitting assumption.

\subsection{Providing Evidence for Predictions}
One critical advantage of the proposed hierarchical pipeline is that it can generate evidence for its predictions at two levels. The first-level identifies a single sentence as the reason for a particular ICD prediction (one sentence \emph{per} predicted ICD code). Moreover, the supervised attention mechanism further points at the exact words in a sentence as the prediction evidence. This property is especially useful when the source sentence is long and contains multiple medical events or concepts. Supervising the attention mechanism with human feedback ensures learning from human coders instead of learning the attention scores unsupervised. 


Figure \ref{fig:case_study_word} exemplifies level two evidence. In this example, we have the following focus sentence and the attention distribution for this sentences' words:

\emph{``Melanocytic Nevus, Lentiginous Junctional Type, With Pigment Incontinence, Completely Excised In The Plane Of Section Examined; Initial And Level Sections.''} 

The human coder assigned D22.71 (Melanocytic nevi of right lower limb, including hip) to this sentence because of the words ``Melanocytic Nevus''. The attention distribution generated by the ICD classifier depicts the same two words receive the highest attention scores.

Typically, attention mechanisms are claimed to provide an interpretation for model predictions \cite{choi2016retain, sha2017interpretable, gao2017hierarchical}. However, in reality, continuous attention scores are usually too close to each other and do not provide much meaningful information. Furthermore, recent studies have shown that attention scores may not correspond to importance scores \cite{jain2019attention, serrano2019attention, sen2020human}. In this work, as we \emph{directly} supervise our attention scores on the annotation data collected from human coders, our attention scores provide meaningful explanations.

\section{Conclusion}
This paper proposes a novel approach for automatic ICD-10 coding by transmuting the extreme multi-label problem into a simpler multi-class problem. We collect phrase-level human coder annotations to supervise our models on learning the specific relations between the input text and predicted ICD codes. Our approach employs two independently trained networks, the sentence tagger and the ICD classifier, stacked hierarchically to predict a clinical report's ICD codeset. The sentence tagger identifies \emph{focus sentences} containing a medical event or concept relevant to an ICD coding. Using a supervised attention mechanism, the ICD classifier then assigns the appropriate ICD code to each focus sentence. To the best of our knowledge, this is the first work to model the ICD prediction task as a multi-class classification problem by breaking it into two sub-tasks. The proposed approach outperforms well-established baseline models by large margins of 23\% in subset accuracy, 18\% in micro-F1, and 15\% in instance-based F-1 using a relatively large test set. Furthermore, our proposed method renders superior interpretability with the help of two-level design and supervised attention. Our main difference compared to other attention-based solutions lies in supervised attention. With the help of attention supervision, interpretability is achieved not through implicitly learned attention scores but by attributing each prediction to a particular sentence and words selected by human coders.

\bibliographystyle{ACM-Reference-Format}
\bibliography{sample-sigconf}

\end{document}